\documentclass[letterpaper, 10 pt, conference]{ieeeconf}  

\IEEEoverridecommandlockouts                              

\usepackage{amsmath, amssymb,bbm,bm,mathtools}
\usepackage[ruled,vlined]{algorithm2e}
\usepackage{comment}
\usepackage{url}

\usepackage{graphicx} 
\usepackage{epsfig} 
\usepackage{amsmath} 
\usepackage{amssymb}  
\usepackage{textcomp}
\usepackage[format=plain,textfont=footnotesize, 
labelfont=footnotesize,
tableposition=top,justification=justified,belowskip=-3pt,aboveskip=8pt]{caption}
\usepackage{xcolor}

\renewcommand{\Re}{\mathbb{R}}
\newcommand{\X}{\mathbf{x}}
\newcommand{\U}{\mathbf{u}}
\newcommand{\V}{\mathbf{v}}

\newcommand{\W}{\mathbf{w}}
\newcommand{\e}{\bm{\epsilon}}

\newcommand{\T}{\intercal}
\newcommand{\E}{\mathbb{E}}
\newcommand{\Prob}{\mathbb{P}}
\newcommand{\Tx}{\Tilde{x}}
\newcommand{\TX}{\Tilde{\mathbf{x}}}

\newcommand{\half}{\mbox{$\frac{1}{2}$}}

\renewcommand{\baselinestretch}{0.98}

\title{\LARGE \bf
Risk-Aware Model Predictive Path Integral Control \\Using Conditional
Value-at-Risk
}

\author{Ji Yin, Zhiyuan Zhang, and Panagiotis Tsiotras
\thanks{The authors are with the D. Guggenheim School of Aerospace Engineering, 
Georgia Institute of Technology, GA. E-mail: {\tt\small \{jyin81,nickzhang,tsiotras\}@gatech.edu}}%
}

\begin{document}

\maketitle
\thispagestyle{empty}
\pagestyle{empty}

\begin{abstract}
    
    In this paper, we present a novel Model Predictive Control method for autonomous robots subject to arbitrary forms of uncertainty. 
    The proposed Risk-Aware Model Predictive Path Integral (RA-MPPI) control utilizes the Conditional Value-at-Risk (CVaR) measure to generate optimal control actions for safety-critical robotic applications. 
    Different from most existing Stochastic MPCs and CVaR optimization methods that linearize the original dynamics and formulate control tasks as convex programs, the proposed method directly uses the original dynamics without restricting the form of the cost functions or the noise. 
    We apply the novel RA-MPPI controller to an autonomous vehicle to perform aggressive driving maneuvers in cluttered environments. 
    Our simulations and experiments show that the proposed RA-MPPI controller can achieve about the same lap time with significantly fewer collisions compared to the baseline MPPI controller. 
    The proposed controller performs on-line computation at an update frequency of up to 80~Hz, utilizing modern Graphics Processing Units (GPUs) to multi-thread the generation of trajectories as well as the CVaR values. 
    \end{abstract}

\section{Introduction}

    Model Predictive Control (MPC) is a control approach that optimizes the control input at the current time while taking the future evolution of the system states and the corresponding controls into account. Compared with the more traditional controllers, MPC controllers can handle complicated interactions between the inputs and outputs while satisfying the given constraints. Moreover, the MPC's ability to predict future events based on the simulated trajectories make it more responsive to dynamic environments than other methods. 
    In recent years, MPC controllers have gained popularity in robotics as massive computational power offered by modern computers allowed the ability to perform fast computations on-the-fly.
    
    MPC methods can be categorized into three types, deterministic MPC (DMPC), stochastic MPC (SMPC) and robust MPC (RMPC), with increasing robustness levels. DMPC assumes that the dynamics are noise-free \cite{ILQR, NMPC}, thus they lack the ability to analyze the uncertainties present in the planning phase and are especially vulnerable to unexpected disturbances. RMPC generates the safest solutions among the three types of controllers because it avoids the worst-case scenarios induced by potential disturbances, albeit at the cost of overly conservative solutions \cite{RMPC}. SMPC can handle dynamical systems with stochastic uncertainty subject to chance constraints \cite{PTMPC,PCMPC, SBMPC1}. SMPC generally produces more cost-efficient solutions compared to RMPC, and achieves more robust planning compared to the DMPC. 
    Although SMPC possesses a superior ability to utilize the probabilistic nature of the uncertainties, most of SMPC approaches suffer from several limitations when it comes to performance guarantees. First, the optimization process is usually designed for a specific form of stochastic noise \cite{CSSMPC}. 
    As a result, robustness is greatly diminished when the environment changes and the assumptions about the form of the noise become invalid. 
    Secondly, many SMPC approaches assume the dynamics to be linear and require linearization of the system during implementation \cite{CSSMPC, Hewing_2018, KOUVARITAKIS20101719}, which slows down computation and imposes strong limitations on these controllers due to the fact that the linearized system can only capture the behavior of the real systems in a small region. 
    Thirdly, the chance constraints used by most stochastic MPC approaches do not account for the seriousness of violating the constraints or the severity of potential accidents. 
    In other words, chance constraints limit the probability of constraint violations, but they fail to distinguish the different costs at stake for violating the constraints.
    
    In this paper, we propose the Risk-aware MPPI (RA-MPPI) algorithm that uses Conditional Value-at-risk (CVaR) \cite{CVaR} to solve the aforementioned problems of existing SMPC methods. 
    The MPPI is a simulation-based nonlinear controller that solves optimal control problems in a receding horizon control setting. Compared with most other MPC methods, MPPI has fewer restrictions on the form of the objective functions and dynamics. 
    Specifically, it can accept non-convex, and even gradient-free costs and nonlinear dynamics. 
    MPPI samples thousands of control sequences by injecting Gaussian noise to a mean control sequence, then it rolls out simulated trajectories following the sampled controls. 
    It then computes the optimal control sequence by taking the weighted average of the costs of the sampled trajectories. Using the parallel computing abilities of modern GPUs, MPPI can achieve on-line planning with a control frequency sufficient for time-critical tasks such as aggressive off-road autonomous driving \cite{CCMPPI}. 
    
Despite its attractive properties, MPPI is still a deterministic control design method, thus it does not account for dynamical noise, not to mention the risks induced by external disturbances. To alleviate this situation, Tube-MPPI~\cite{TubeMPPI} uses iLQG as an ancillary controller to track the nominal trajectory generated by the original MPPI. 
However, Tube-MPPI does not allow explicit assignment of risk levels during planning, making it difficult to specify the desired trade-off between risk and robustness. 
As a remedy, Robust MPPI (RMPPI)~\cite{RobustMPPI} improves upon Tube-MPPI by introducing an upper bound on the free-energy growth of the system to describe a task constraint satisfaction level, however, the free-energy bound does not provide an intuitive interpretation of risk. Moreover, both Tube-MPPI and RMPPI use the linearized system dynamics to solve another ancillary optimization problem, which is problematic because linearizing the original dynamics is time-costly for real-time applications and is only accurate within a small region. 
RMPPI can alleviate this problem by replacing iLQG with Control Contraction Metric (CCM) \cite{CCMs}, which is a feedback controller that provides exponential convergence guarantee for nonlinear systems \cite{RobustMPPI}. However, CCM cannot be computed in real-time either. On the other hand, CVaR is a risk metric that has gained popularity in many engineering problems recently, due to its appealing ability to measure and prevent outcomes that hurt the most \cite{VaR}. 
However, most existing CVaR optimization algorithms share the same problem with the current SMPC approaches, namely, they normally assume convex objective functions and constraints\cite{Krok, ahmadi2021risk}, thereby limiting their applicability. Other approaches utilize dynamic programming to optimize time-consistent modifications of CVaR \cite{chow2014framework, ruszczynski2010risk, artzner2007coherent}, which requires careful interpretation of their practical meanings. 
    
Overall, the main contribution of this paper is threefold: 
First,  we propose the novel RA-MPPI controller, which improves the baseline MPPI controller by making it account for the control risks in terms of CVaR. 
The novel RA-MPPI controller neither limits the form of the state-dependent costs nor restricts the type of system noise. 
Second, the proposed RA-MPPI utilizes a novel, general control architecture that can be easily modified to adapt various risk metrics other than CVaR. 
For example, it can be integrated with chance constraints \cite{CSvehicle} or some other time-consistent risk metric \cite{Samuelson2018SafetyAwareOC, MarcoPavoneCVaR, RATMM}.
Third,
we demonstrate the proposed controller by testing it on challenging autonomous racing tasks in cluttered environments. Our simulations and experiments show that the proposed RA-MPPI runs on-line at about 80~Hz using a dynamic vehicle model with tire dynamics, and it outperforms the MPPI controller in terms of collisions while maintaining approximately the same lap time, given various assigned risk levels.

\section{MPPI Review}\label{Problem Formulation}

Consider a general discrete nonlinear dynamical system,
\begin{equation}\label{dynamics}
    x_{k+1} = F(x_k, u_k),
\end{equation}
where $x_k \in \Re^{n_x}$ is the system state and $u_k \in \Re^{n_u}$ is the control input. Within a control horizon $K$, we denote the trajectory $\X = \left[x_0^\T,\ldots,x_K^\T\right]^\intercal \in \Re^{n_x (K+1)}$, the mean control sequence $\V = \left[v_0^\T,\ldots,v_{K-1}^\T\right]^\intercal \in \Re^{n_u K}$, the injected Gaussian control noise $\e = \left[\epsilon_0^\T,\ldots,\epsilon_{K-1}^\T\right]^\intercal \in \Re^{n_u K}$, and the disturbed control sequence $\U = \V + \e$.

The original Model Predictive Path Integral control (MPPI) solves the following problem, 
\begin{subequations}\label{Problemformulation}
\begin{align}
    \min_\V J(\V) &= \nonumber \\
    &\mathbb{E}\left[ \phi(x_K) + \sum^{K-1}_{k=0} \left(q(x_k)  + \frac{\lambda}{2}v_k^\T \Sigma^{-1}_\epsilon v_k \right) \right],\label{MPPI objective}
\end{align}
subject to
\begin{align}
&x_{k+1} = F(x_k, v_k + \epsilon_k),\label{NonlinearSystem}\\
&x_0 = x(0), \quad \epsilon_k \sim \mathcal{N}(0, \Sigma_{\epsilon}), \label{InitialCondition}
\end{align}
\end{subequations}
where $x(0)$ is the measured current state. Note that the state-dependent cost $q(x_k)$ can take an arbitrary form. 
The MPPI controller is derived by minimizing the KL-divergence between the current controlled trajectory distribution and the optimal distribution \cite{TubeMPPI} to solve the problem \eqref{Problemformulation}. 
It samples a large number of simulated trajectories to synthesize the optimal control sequence during each optimization iteration. Assuming MPPI samples trajectories $m = 0,\hdots, M$ in simulation at each iteration, the cost $S_m$ of the $m^{th}$ sample trajectory is evaluated as,
\begin{equation}\label{TrajectoryCost}
    S_m = \phi(x^{m}_K) + \sum^{K-1}_{k=0}q(x^{m}_k) +
     \gamma v_k^\intercal \Sigma_\epsilon^{-1} (v^{m}_k + \epsilon^{m}_k),  
\end{equation}
where $x^m_k \in \Re^{n_x}$ is the system state, $\epsilon^m_k \in \Re^{n_u}$ is the control noise of the $m^{th}$ sampled trajectory, and $\gamma \in \left[0, \lambda \right]$ is the weight for the control costs.  The MPPI algorithm calculates the optimal control $\V^+$ by taking the weighted average of all sampled control sequences,
\begin{equation}\label{OptimalControl}
\V^+ = \sum^{M}_{m=1}\omega_m \U^{m}/ \sum^{M}_{m = 1}\omega_m,
\end{equation}
where $\U^m = \V + \e^m$ is the control sequence corresponding to the $m^{th}$ simulated trajectory. The corresponding weight $w_m$ for $\U^m$ is determined by, 
\begin{equation}\label{Weights}
    \omega_m = \text{exp}\left(-\frac{1}{\lambda}\left(S_m - \beta \right)\right),
\end{equation}
where
$
    \beta = \min_{m=1,\ldots,M} S_m,
$
is the smallest trajectory cost among the $M$ sampled trajectories, and it is used to prevent numerical overflow while keeping the solution the same \cite{RobustMPPI}. 
The mean control $\V$ for the next receding horizon control iteration is then set to be $\V = \V^+$.

\section{Risk-aware MPPI}\label{RA-MPPI}
\begin{figure}[t]
\centerline{\includegraphics[scale = 0.27]{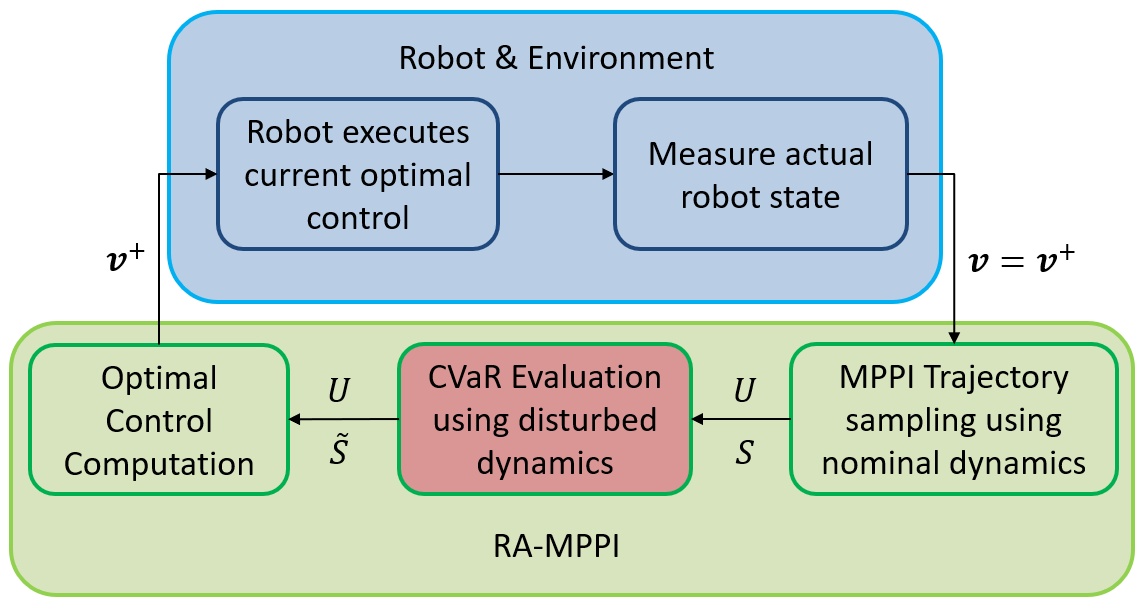}}
\caption{The proposed RA-MPPI control architecture. The set $U = \{\U^0,\hdots,\U^M \}$ contains all the sampled control sequences, and the set $S = \{S_0, \hdots, S_M \}$ includes their corresponding trajectory costs. The set $\Tilde{S} = \{\Tilde{S}_0, \hdots, \Tilde{S}_M\}$ contains the modified trajectory costs which are obtained by adding the CVaR constraint violation costs evaluated using the disturbed dynamics \eqref{disturbed_dynamics} to the original MPPI sampled trajectory costs $S$. }
\label{RA-MPPI Schematics}
\end{figure}

The MPPI problem formulation in \eqref{Problemformulation} only minimizes the cost for the expected performance of a system injected with some Gaussian control noise $\epsilon_k$, and it follows a deterministic dynamical system \eqref{dynamics} without considering risk explicitly in the loop. 
To take risks into account, we assume that the simulated dynamics \eqref{dynamics} is an unbiased estimate of a more general, stochastic system,
\begin{equation}\label{disturbed_dynamics}
    \Tilde{x}_{k+1} = \Tilde{F}(\Tilde{x}_k, u_k, w_k),\quad \Tilde{x}_0 = x_0,
\end{equation}
such that,
\begin{equation}\label{nominal_dynamics}
    F(x_k, u_k) = \E[\Tilde{F}(\Tilde{x}_k, u_k, w_k)],
\end{equation}
 where $\Tilde{\X} = [\Tilde{x}_0^\T,\hdots,\Tilde{x}_K^\T]^\T \in \Re^{n_x(K+1)}$ is the trajectory realization following the disturbed dynamics \eqref{disturbed_dynamics}. In this work, we are interested in minimizing the objective function \eqref{MPPI objective} while preventing worst-case, catastrophic scenarios. To this end, the proposed RA-MPPI solves the following problem,
\begin{subequations} \label{RAMPPI_Problem_Formulation}
\begin{align}
\min_\V J(\V) \text{ subject to } \eqref{NonlinearSystem}, \eqref{InitialCondition} \text{ and,}\\
\textrm{CVaR}_{\alpha}(L(\Tilde{\X})) \leq C_{u}, \label{CVaR_constraint}\\
\Tilde{x}_{k+1} = \Tilde{F}(\Tilde{x}_k, v_k, w_k), \quad \Tilde{x}_0 = x_0, \label{initialcondition}
\end{align}
\end{subequations}
where the parameter $\alpha \in (0,1)$ is the CVaR confidence level, and $C_{u}$ determines the upper bound of the CVaR constraint of the state-dependent risk cost $L(\Tilde{\X})$ defined by \eqref{TotalCollisionCost} in Section \ref{CVaREvaluationSection}. Figure \ref{RA-MPPI Schematics} illustrates the RA-MPPI control architecture. Section \ref{CVaREvaluationSection} discusses the evaluation of CVaR, and Section \ref{CVaRfiltering} introduces the details of applying CVaR to filter sampled trajectories such that the proposed RA-MPPI generates risk-aware controls satisfying the constraint \eqref{CVaR_constraint}.

\subsection{Monte-Carlo CVaR Evaluation}\label{CVaREvaluationSection}
We define the risk cost $L(\TX)$ in \eqref{CVaR_constraint} as,
\begin{equation}\label{TotalCollisionCost}
    L(\TX) = \sum^{K-1}_{k=0}l(\Tx_k),
\end{equation}
where $l(\Tx_k)$ is the running cost of arbitrary form. The Conditional Value-at-Risk is defined as,
\begin{equation}\label{CVaR}
    \textrm{CVaR}_\alpha(L(\TX)) = \E[L(\TX) \,| \, L(\TX) \geq \textrm{VaR}_\alpha(L(\TX))],
\end{equation}
where the value at risk (VaR) is defined as \cite{VaR},
\begin{equation}\label{VaR}
    \textrm{VaR}_\alpha(L(\TX)) = \min \{ t \, |\, \Prob(L(\TX) \leq t) \geq \alpha\},
\end{equation}
that is, the risk cost $L(\TX)$ has a probability of $1 - \alpha$ to be greater than $\textrm{VaR}_\alpha(L(\TX))$. It follows that the CVaR value \eqref{CVaR} can be interpreted as the expected risk cost of the trajectory $\TX$ among the worst $1-\alpha$ quantile of the cost distribution $L(\TX)$. In addition to sampling $M$ noise-free trajectories using the nominal system \eqref{dynamics} as the original MPPI, the proposed RA-MPPI controller quantifies the risk of the $m^{th}$ noise-free trajectory by simulating another $N$ trajectories following the disturbed system \eqref{disturbed_dynamics}, for $m = 0,\hdots,M-1$. For simplicity, we drop the brackets in \eqref{CVaR}, \eqref{VaR}, and use $\textrm{CVaR}^m_\alpha$, $\textrm{VaR}_\alpha^m$ to denote the CVaR and VaR values for the $m^{th}$ noise-free trajectory. 
To this end, $\textrm{CVaR}^m_\alpha$ can be approximated from,
\begin{equation}\label{CVaREvaluationEquation}
    \textrm{CVaR}_\alpha^m = \frac{1}{N_o}\sum^{N-1}_{n=0}\mathbf{1}_{\textrm{CVaR}_\alpha}(L(\TX^{m,n}))\cdot L(\TX^{m,n}),
\end{equation}
where the simulated trajectories $X^m = \{\TX^{m, 0},\hdots,\TX^{m,N-1}\}$ are sampled using the $m^{th}$ sampled control sequence $\U^m$ and following the disturbed dynamics \eqref{disturbed_dynamics}, given the initial state $\Tx^{m, n}_0 = x_0$. 
In \eqref{CVaREvaluationEquation}, $N_o$ is the number of trajectories among $X^m$ with cost $L(\TX^{m,n}) \geq \textrm{VaR}_\alpha^m$, and the indicator function is defined as,
\begin{equation}
    \mathbf{1}_{\textrm{CVaR}_\alpha}(L(\TX^{m,n}))=\left\{
            \begin{array}{ll}
              1, \quad \textrm{if } L(\TX^{m,n}) \geq \textrm{VaR}_{\alpha}^m,\\
              0, \quad \textrm{otherwise}.
            \end{array}
          \right.
\end{equation}
In case $\alpha$ approaches 1, using the method in \cite{deo2021efficient} can improve the sampling efficiency for the CVaR evaluation. 
We do not consider this special case in this paper.
The proposed control architecture is illustrated in Figure \ref{RA-MPPI Schematics}. 
Note that the red block in the figure evaluates the CVaR risk metric using the Monte-Carlo sampling method, and thus can be easily integrated with other risk metrics such as chance constraints.

\subsection{Soft Trajectory Filtering using CVaR}\label{CVaRfiltering}

The RA-MPPI filters the MPPI sampled trajectories by adding their corresponding CVaR values as penalty costs to the original MPPI trajectory costs evaluated by \eqref{TrajectoryCost}, thus penalizing MPPI sampled trajectories with high risk. 
Let us define the CVaR constraint violation cost $J^m_C$ for the $m^{th}$ noise-free trajectory,
\begin{equation}\label{CVaRCost}
     J_{C}^m = A\cdot \textrm{CVaR}_{\alpha}^m\cdot \mathbf{1}_{C_{u}}(\textrm{CVaR}_{\alpha}^m),
\end{equation}
where,
\begin{equation}\label{JC}
    \mathbf{1}_{C_{u}}(\textrm{CVaR}_{\alpha}^m)=\left\{
    \begin{array}{ll}
      1, \quad \textrm{if } \textrm{CVaR}_{\alpha}^m > C_{u},\\
      0, \quad \textrm{otherwise}.
    \end{array}
  \right.
\end{equation}
Note that the coefficient $A$ in \eqref{CVaRCost} adjusts the magnitude of the CVaR cost, and the parameter $C_u$ in \eqref{JC} determines how strong the CVaR constraint \eqref{CVaR_constraint} is. Adding \eqref{CVaRCost} to \eqref{TrajectoryCost}, we obtain the modified cost for the $m^{th}$ trajectory as,
\begin{equation}\label{TrajectoryCostModified}
    \Tilde{S}_m = J^m_C + \phi(x^{m}_K) + \sum^{K-1}_{k=0}q(x^{m}_k) + \gamma {v_k}^\T \Sigma_\epsilon^{-1}u^{m}_k.
\end{equation}
It follows from \eqref{Weights} that the updated weight for the $m^{th}$ sampled control sequence $\U^m$ is,
\begin{equation}\label{WeightComputation}
    \Tilde{\omega}_m = \text{exp}\left(-\frac{1}{\lambda}\left(\Tilde{S}_m - \Tilde{\beta} \right)\right),
\end{equation}
where $\Tilde{\beta} = \min_{m=1,\ldots,M} \Tilde{S}_m$ and the resulting optimal control sequence can be obtained by using \eqref{OptimalControl}.

\subsection{Risk Cost Sensitivity Scaling}

In case $J^m_C$ is relatively small compared to $\Tilde{S}_m$ in \eqref{TrajectoryCostModified}, a change in the $\textrm{CVaR}^m_\alpha$ value in \eqref{CVaRCost} has limited impact on the relative change of $\Tilde{S}_m$, thus the resulting optimal control could become insensitive to risk variations. Increasing the coefficient $A$ in \eqref{CVaRCost} can potentially alleviate the situation, however, adjusting only the value of $A$ may let $J^m_C$ dominate \eqref{TrajectoryCostModified}, leading to overly conservative solutions. A simple remedy to this problem is to scale up the variance of the risk costs while maintaining their mean value. To this end, given a set of risk costs $L^m = \{L(\TX^{m,0}), \hdots, L(\TX^{m,N-1})\}$ in \eqref{CVaREvaluationEquation} for the $m^{th}$ noise-free trajectory, we can obtain the updated risk costs with adjusted variance by,
\begin{equation}\label{AdjustedRiskCost}
    L_a(\TX^{m,n}) = B \cdot (L(\TX^{m,n}) - \bar{L}^m) + \bar{L}^m,
\end{equation}
where $\bar{L}^m$ is the mean of the original set $L^m$, given by,
\begin{equation}\label{OriginalCostMean}
    \bar{L}^m = \sum^{N-1}_{n=0} L(\TX^{m,n}),
\end{equation}
and where $B$ is a scaling factor. 
Increasing $B$ will make the proposed RA-MPPI more sensitive to risk level changes. 
We can then use the adjusted risk costs $L^m_a$ instead of $L^m$ in \eqref{CVaREvaluationEquation} to evaluate $\textrm{CVaR}^m_\alpha$.

\section{The Risk-Aware MPPI Algorithm}\label{Risk-aware MPPI Algorithm}
In this section, we introduce the RA-MPPI algorithm. In Algorithm \ref{algo1},  line \ref{GetEstimateLine} gets the current estimate of the system state $x_0$. Line \ref{MPPITrajectoriesRolloutBeginLine} to \ref{MPPITrajectoriesRolloutEndLine} samples the control sequences, rolls out and evaluate the simulated trajectories following the deterministic system \eqref{dynamics}. Line \ref{CVARBeginLine} to line \ref{CVAREndLine} samples the simulated trajectories used for evaluating the CVaR values following the disturbed dynamics \eqref{disturbed_dynamics}, then line \ref{TrajectoryFilteringBeginLine} to line \ref{TrajectoryFilteringEndLine} compute CVaR values and carry out trajectory filtering. More specifically, line \ref{WarmControl} produce sampled controls by adding control noises to mean controls. Line \ref{ZeroMeanControl} introduces some zero-mean controls to help smooth the optimal controls \cite{CCMPPI}. Line \ref{TrajectoryFilteringBeginLine} scales up the sensitivity of the risk costs following \eqref{AdjustedRiskCost} and \eqref{OriginalCostMean}. Line \ref{FindWorstQuantileLine} finds the set $P$ that contains the upper $1-\alpha$ quantile of the risk cost set $\{L(\TX^{m,0}),\hdots,L(\TX^{m,N-1})\}$, and line \ref{AverageP} calculates the CVaR using \eqref{CVaREvaluationEquation}. Line \ref{UpdateTrajectoryCost} adds the CVaR constraint \eqref{CVaR_constraint} violation costs to the trajectory costs $S_m$ to obtain the modified trajectory costs $\Tilde{S}_m$ using \eqref{CVaRCost} and \eqref{TrajectoryCostModified}. Based on the trajectories $\Tilde{S} = \{S_0,\hdots,S_M \}$ and the control sequence set $U = \{\U^0,\hdots,\U^M \}$, line \ref{CalculateOptimalControlLine} computes the optimal control sequence $\V^+$ using \eqref{Weights} and \eqref{OptimalControl}. Line \ref{ExecuteCommandLine} sends the first control $v_0^+$ to the actuators, and line \ref{InitializationLine} sets the optimal controls to be the mean controls for the next optimization iteration.

\begin{algorithm}[h]
    \caption{Risk-Aware MPPI Algorithm}\label{algo1}
    \SetAlgoLined
    \LinesNumbered
    \SetKwInOut{Input}{Given}
    \Input{
    \noindent $\text{RA-MPPI Parameters } \gamma, \eta, \alpha, A, B, C_u$;}
    \SetKwInOut{Input}{Input}
    \Input{
    \noindent $\text{Initial control sequence } \V$}
    
    \While{task not complete}{
        $x_{0} \leftarrow \textit{GetStateEstimate}()$;\\ \label{GetEstimateLine}
        \For{$m\leftarrow 0 \textbf{ to } M-1$ in parallel}{\label{TrajectorySamplingBeginLine} 
              $x^m_0 \leftarrow x_0, \quad S_m \leftarrow 0$;\label{MPPITrajectoriesRolloutBeginLine}\\
              $\text{Sample }\e^{m} \leftarrow \{\epsilon^{m}_0,\ldots,\epsilon^{m}_{N-1}\}$;\\
              \For{$k\leftarrow 0 \textbf{ to } K-1$}{
                      \uIf{$m < (1-\eta)M$}{ \label{ControlandUncontrolledCommandsBeginLine}
                      $u^{m}_{k} \leftarrow v_{k} + \epsilon^{m}_k$;\label{WarmControl}
                      }\Else{
                      $u^{m}_{k} \leftarrow \epsilon^{m}_k$;\label{ZeroMeanControl}\\ }\label{ControlandUncontrolledCommandsEndLine}
                      $x^{m}_{k+1} \leftarrow F(x^{m}_k , u^{m}_k)$;\\
                      $S_m \leftarrow S_m + q_k(x^{m}_k) + \gamma {v_k}^\T \Sigma_\epsilon^{-1}u^{m}_k$;\\
                      }
                $S_m \leftarrow S_m + \phi(x^{m}_N)$;\\ \label{MPPITrajectoriesRolloutEndLine}
                \For{$n\leftarrow 0 \textbf{ to } N-1$ in parallel}{\label{CVARBeginLine}
                $\Tx^{m,n}_{0}\leftarrow x_0, \quad L(\TX^{m,n}) \leftarrow 0;$\\
                $\text{Sample }\W^{m,n} \leftarrow \{w^{m,n}_0,\ldots,w^{m,n}_{N-1}\}^\intercal$;\\
                  \For{$k\leftarrow 0 \textbf{ to } K-1$}{
                      $\Tx^{m,n}_{k+1} \leftarrow \Tilde{F}(\Tx^{m,n}_k , u^{m}_k, w^{m,n}_k)$;\\
                      $L(\TX^{m,n}) \leftarrow L(\TX^{m,n}) + l(\Tx_k^{m,n})$;\\
                      }
                }\label{CVAREndLine}
                $L^m_a \leftarrow Scal(\{L(\TX^{m,0}),\hdots,L(\TX^{m,N-1})\},B)$;\\\label{TrajectoryFilteringBeginLine}
                $P \leftarrow Max(L^m_a, 1-\alpha)$; \\ \label{FindWorstQuantileLine}
                $\textrm{CVaR}_\alpha^m = Average(P)$;\label{AverageP}\\
                \If{$\textrm{CVaR}_\alpha^m > C_{u}$}
                {
                    $\Tilde{S}_m \leftarrow S_m + A\cdot CVaR_{\alpha}^m$;\label{UpdateTrajectoryCost}\\
                }\label{TrajectoryFilteringEndLine}
        } \label{TrajectorySamplingEndLine}
        $\V^+ = \textit{CalculateOptimalControl}(\Tilde{S}, U)$;\\\label{CalculateOptimalControlLine} 
        $\textit{ExecuteCommand}(v_{0}^+)$;\\ \label{ExecuteCommandLine} 
        $\V = \V^+$;\label{InitializationLine}
    }
\end{algorithm}

\section{Simulation and Experiment}\label{Numerical Simulations Section}

In this section, we demonstrate the proposed RA-MPPI controller and compare it with the baseline MPPI controller on an autonomous racing platform. We show that the novel RA-MPPI significantly improves robustness against disturbances in challenging autonomous driving tasks by using a variety of simulation and experiment examples. 
All the simulation and experiment results are obtained by running the controllers in real time. 

\subsection{Controller Setup}

\begin{figure}[!h]
\centerline{\includegraphics[scale = 0.45]{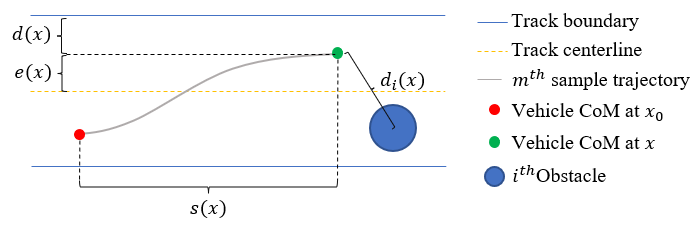}}
\caption{Track schematic.}
\label{TrackSchematic}
\end{figure}

In our simulations, both the proposed RA-MPPI controller and the baseline MPPI controller evaluate the costs of their sampled trajectories, using \eqref{TrajectoryCostModified} and \eqref{TrajectoryCost} respectively. For both controllers,  we use the following running cost,
 \begin{equation}\label{state_dependent_cost}
     q(x)= c_1 \mu_{\textrm{bdry}}(x) + c_2 \mu_{\textrm{obs}}(x) + c_3 q_\textrm{dev}(x),
 \end{equation}
where in the simulations the weights are $c_1 = 2, c_2 = 1, c_3 = 0.1$, and the cost $q_\textrm{dev}(x) = e(x)^2$ penalizes the lateral deviation $e(x)$ between the vehicle's CoM and the track centerline. The differentiable boundary cost $\mu_{\textrm{bdry}}(x)$ is, 
 \begin{equation}\label{BoundaryCost}
     \mu_{\textrm{bdry}}(x)=  \max \Big\{ 0,\frac{\tan^{-1}(-100 d(x))}{\pi} + \half \Big\},\\
 \end{equation}
 and the discrete obstacle cost $\mu_{\textrm{obs}}(x)$ is given by,
  \begin{equation}\label{ObstacleCostDiscontinuous}
     \mu_{i,\textrm{obs}}(x)=\left\{
     \begin{array}{ll}
       1, \quad \textrm{if } d_i(x) < r_i,\\
       0, \quad \textrm{otherwise}.
     \end{array}
   \right.
 \end{equation}
 Notice that \eqref{BoundaryCost} is a smooth approximation of the unit step function that penalizes collisions with track boundaries, and $d(x)$ is the signed distance from the nearest track boundary to the vehicle's CoM, which is a positive number if the vehicle stays on the track. The term $d_i(x)$ in \eqref{ObstacleCostDiscontinuous} measures the distance between the vehicle's CoM to the center of the $i\textrm{th}$ circular obstacle with radius $r_i$. The terminal costs in both \eqref{TrajectoryCost} and \eqref{TrajectoryCostModified} are set the same as,
  \begin{equation}\label{terminal_cost}
     \phi(x) = c_4 - c_5s(x),
 \end{equation}
 where we use the weights $c_4 = 0.6$ and $c_5 = 2$ such that $\phi(x) \geq 0$ for all $x$ in the simulations, and $s(x)$ is the distance traveled by the $m^{th}$ noise-free sampled trajectory by the end of the control horizon along the track centerline. 
 Figure \ref{TrackSchematic} demonstrates the physical meanings of the variables $e(x), s(x), d(x), d_i(x)$ in \eqref{state_dependent_cost}-\eqref{terminal_cost} with a track schematic.
 
 For both the RA-MPPI and the MPPI controllers, we set the control horizon to $K = 30$, the inverse temperature $\lambda$ in \eqref{Weights} and \eqref{WeightComputation} to be $\lambda = 0.35$, and the portion of the zero-mean control trajectories $\eta$ in Algorithm \ref{algo1}, Line \ref{ControlandUncontrolledCommandsBeginLine} to be $\eta = 0.2$.
 For the proposed RA-MPPI controller, we set the number of sampled noise-free trajectories to be $M = 1024$, and for each such trajectory, we sample $N = 300$ to evaluate its corresponding CVaR value using \eqref{CVaREvaluationEquation}. 
 As a result, the RA-MPPI samples $3.072\times 10^5$ trajectories in total at each time step. 
 We further let the weight $A$ of the CVaR constraint violation cost in \eqref{CVaRCost} be $A = 10$, and the running risk cost $l(\Tilde{x}_k)$ in \eqref{TotalCollisionCost} be the same as the state-dependent cost \eqref{state_dependent_cost}, such that $l(\Tilde{x}_k) = q(\Tilde{x}_k)$. 
 The number of sampled trajectories for the baseline MPPI is set to be equal to the total number of sampled trajectories of the RA-MPPI, in order to make fair comparisons between the controllers.

\subsection{Aggressive Driving in Cluttered Environments Subject to Disturbances}\label{SimulationSession}

\begin{figure}[!h]
\centerline{\includegraphics[scale = 0.40]{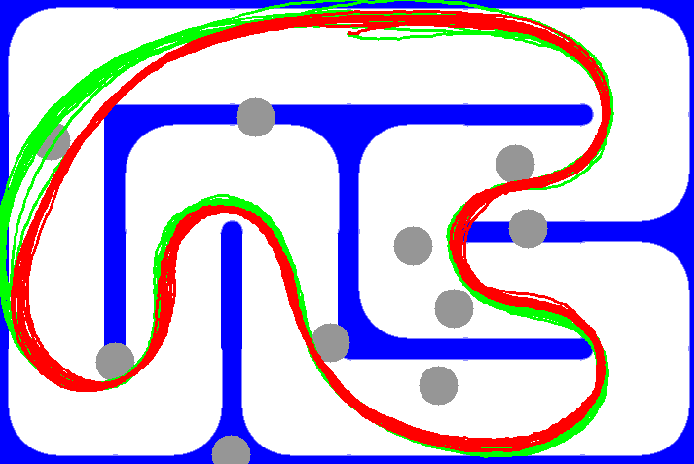}}
\caption{Samples of RA-MPPI and MPPI trajectories on the race track. 
The trajectories in red are generated by the proposed RA-MPPI controller, and the trajectories in green are generated by the baseline MPPI controller.}
\label{track_RAMPPI_MPPI}
\end{figure}

Our simulations use a 0.6~m wide track with a centerline of length 10.9~m, and each corner of the track has a turning radius 0.3~m. There are 10 circular obstacles of the same radius $r_i = 0.1$~m scattered randomly on the track.
The track is shown in  Figure~\ref{track_RAMPPI_MPPI}.
In addition, we use a combination of the discrete obstacle cost \eqref{ObstacleCostDiscontinuous} and the continuous boundary cost \eqref{BoundaryCost} for the track to test the robustness of the controllers, and the controllers are set up to drive the vehicle as fast as possible. 

We run both the proposed RA-MPPI and the original MPPI controllers with system  \eqref{disturbed_dynamics} using Gaussian noise $w_k \sim \mathcal{N}(0, 0.2I)$, in order to compare their performance in the presence of dynamical disturbances. 
Each controller runs the vehicle for 100 laps and Figure~\ref{track_RAMPPI_MPPI} shows the resulting trajectories. 

The RA-MPPI trajectories show strong awareness in avoiding collisions in this aggressive driving setting, resulting in $80\%$ fewer collisions than the MPPI controller, while achieving approximately the same lap time. 
This result indicates that the RA-MPPI can prevent the worst-case scenarios without sacrificing the average performance of the controller. 

\begin{figure}[!h]
\centerline{\includegraphics[scale = 0.45]{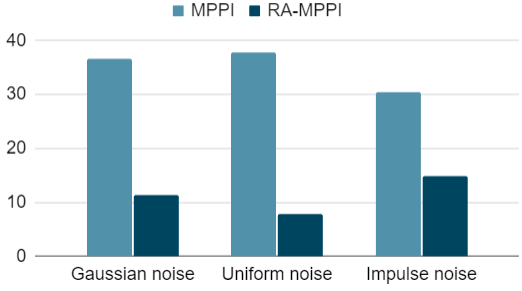}}
\caption{The average number of collisions per lap for the RA-MPPI and MPPI controllers subject to various forms of  noise. 
The RA-MPPI controller outperforms the MPPI controller by reducing the number of collisions by at least $50\%$ for all of the three selected types of noise.}
\label{NoiseVariationsPlot}
\end{figure}

As the proposed control architecture in Figure \ref{RA-MPPI Schematics} utilizes Monte-Carlo sampling to evaluate CVaR, the RA-MPPI controller accepts arbitrary forms of disturbances $w_k$ in \eqref{disturbed_dynamics}. To this end, we examined the robustness of the proposed RA-MPPI controller by repeating the simulations in 
Figure~\ref{track_RAMPPI_MPPI} using various forms of $w_k$. 
Specifically, in addition to the normally distributed disturbance $w_k \sim \mathcal{N}(0, 0.2I)$, we also use uniformly distributed noise $w_k \sim \mathcal{U}_{[-0.2, 0.2]}$ and a form of impulse/jump noise disturbance $w_k$ that has a $2\%$ chance to give the system state $\Tilde{x}_{k+1}$ in \eqref{disturbed_dynamics} a sudden change of magnitude $0.45$ in arbitrary directions. 
We define a collision to be the situation where the vehicle state is inside some infeasible region, such as the space occupied by obstacles or areas outside of the track. 
The simulation results are shown in Figure \ref{NoiseVariationsPlot}.

\begin{figure}[!h]
\centerline{\includegraphics[scale = 0.35]{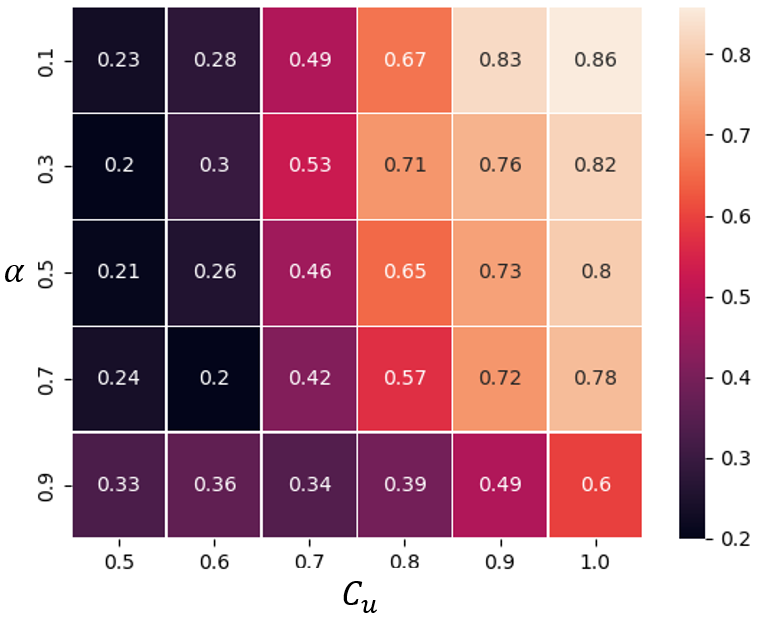}}
\caption{RA-MPPI and MPPI performance comparison. The numbers in this chart shows the performance ratio $(N_{R}/N_{M})$, where $N_R$ is the number of collisions of the proposed RA-MPPI, and $N_M$ is the number of collisions of the MPPI controller. Each data point is collected by running 100 laps.}
\label{TemperaturePlot}
\end{figure}

To further investigate the performance of the proposed RA-MPPI controller with different user-specified risk levels, we performed a grid search by varying the CVaR constraint upper bound $C_u$, and the CVaR confidence level $\alpha$ in \eqref{CVaR_constraint}. 
The results are summarized in Figure~\ref{TemperaturePlot}.  
In the figure, the proposed RA-MPPI algorithm tends to achieve fewer collisions as the CVaR confidence level $\alpha$ increases, despite some  outliers resulting from the randomness of the simulations. 
Moreover, as we tighten the CVaR constraint by decreasing the constraint upper bound value $C_u$, the RA-MPPI shows more robust performance and experience fewer collisions. 
Due to the fact that the CVaR is a time-inconsistent risk metric~\cite{MarcoPavoneCVaR}, applying an overly strict CVaR constraint such as setting $C_u = 0.5$ and $\alpha = 0.9$ in \eqref{CVaR_constraint} puts a large weight on preventing risk evaluated at the current time while aggravating the overall performance~\cite{MarcoPavoneCVaR}. 
To alleviate this situation, we can simply loosen the constraint by using $C_u = 0.6$ and $\alpha = 0.7$, or use a time-consistent modification of CVaR.   


\subsection{Autonomous Racing Experiment}\label{ExperimentSession}

\begin{figure}[!h]
\centering
\includegraphics[width=0.8\linewidth]{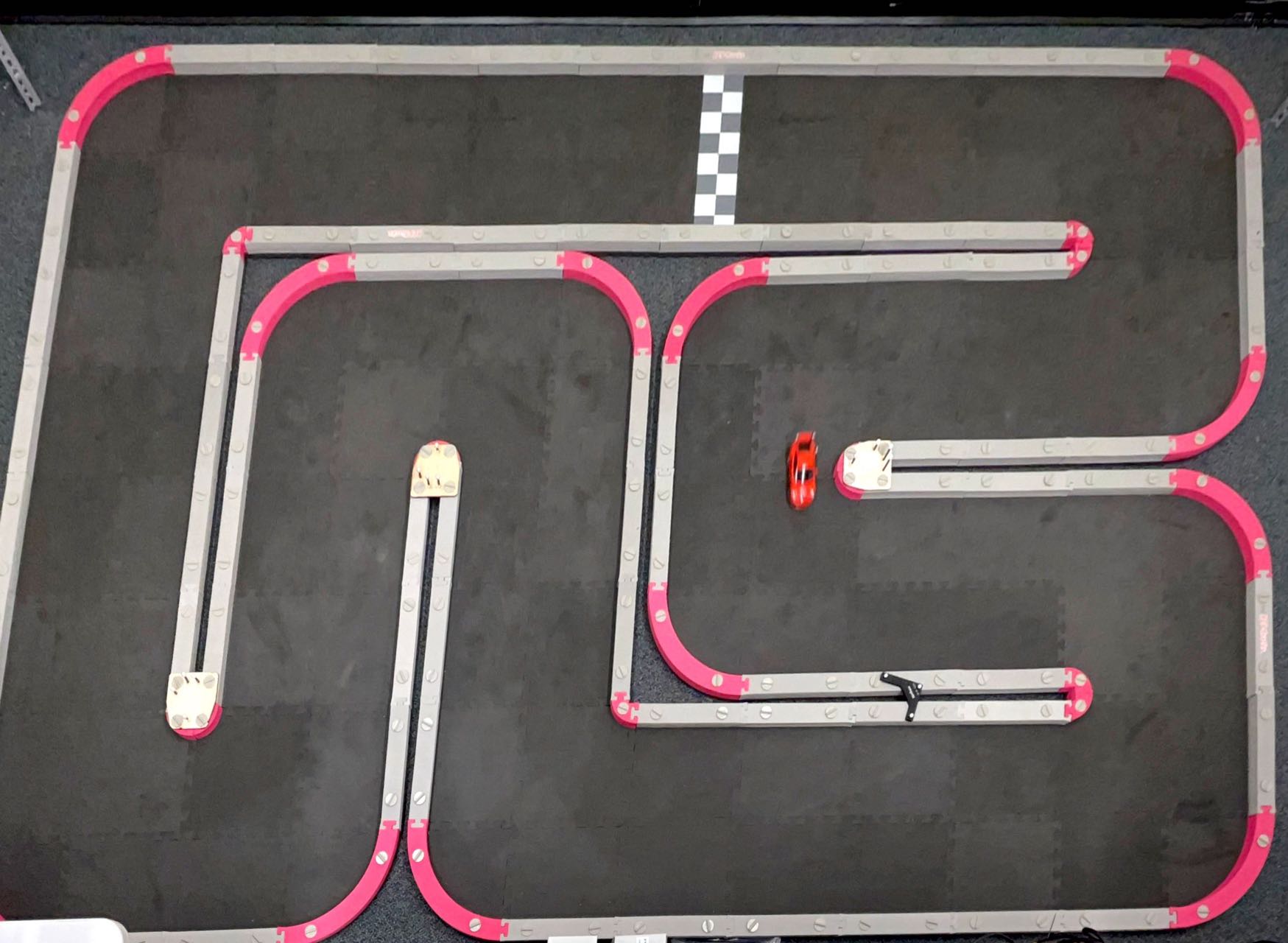}
\caption{The Autonomous vehicle racing on the track.}
\label{Autonomous racing experiment}
\end{figure}

\begin{figure}[!h]
\centerline{\includegraphics[scale = 0.24]{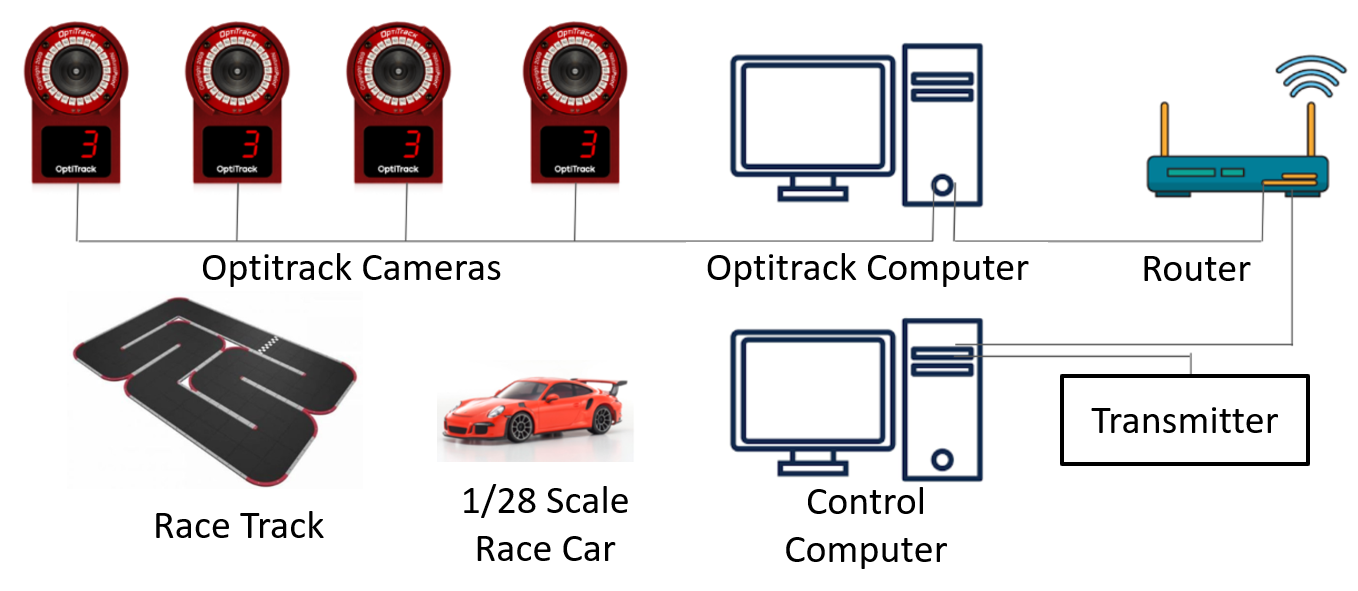}}
\caption{Autonomous racing platform system schematic.}
\label{BuzzracerSystem}
\end{figure}

To validate the proposed RA-MPPI controller in the real world, we implemented the algorithm on an autonomous racing platform and tested its performance.
Our experiment uses the same track layout as the simulations described in Section~\ref{SimulationSession}. 
Figure~\ref{Autonomous racing experiment} shows a snapshot of our autonomous vehicle running on a real-world track. 
A schematic for this autonomous racing system setup is illustrated in Figure~\ref{BuzzracerSystem}.

The system uses ten Optitrack cameras to observe the race track and the motion of a 1/28 scale race car on the track. 
These cameras pass raw data to the Optitrack computer, which then computes the state of the race car. 
 The system utilizes the Optitrack visual tracking system for state updates of the vehicle. 
The state measurements are broadcasted on a local network to the control computer,
In our experiment, we run each controller for 10 laps. 
The collected results are shown in Table~\ref{ExperimentTable}.
These results show that the RA-MPPI experiences $55.7\%$ fewer collisions with approximately the same lap time.

\begin{table}[!h]
\caption{RA-MPPI and MPPI experimental results.}
\centering
\begin{tabular}{|c|c|c|}
\hline
\textbf{Controller Type} & \textbf{No.Collision/lap} & \textbf{Avg. Laptime(s)} \\
\hline
$\textrm{MPPI}$      & 20.3 & 7.11 \\
\hline
$\textrm{RA-MPPI}$      & 9.0 & 7.38 \\
\hline
\end{tabular}
\label{ExperimentTable}
\end{table}

The performance of the proposed RA-MPPI Algorithm~\ref{algo1} requires a parallel implementation of the control architecture 
described in Figure~\ref{RA-MPPI Schematics}. 
Compared to our previous implementation of MPPI in \cite{CCMPPI}, we reduced the data exchange between the GPU and CPU to a minimum by moving all necessary computations to GPU. 
Specifically, our implementation of the RA-MPPI controller computes the dynamics propagation, including both the nominal and the disturbed dynamics, and evaluates CVaR and the trajectory costs in parallel on the GPU. 
On average, the resulting RA-MPPI can sample and evaluate $1.024\times 10^5$ trajectories at 81.6~Hz, $3.072\times 10^5$ trajectories at 44.3~Hz and $5.120\times 10^5$ trajectories at 27.8~Hz, using an Nvidia GeForce RTX3090 GPU.

\section{Conclusions And Future Work}
In this paper, we proposed the RA-MPPI, a risk-aware model predictive control scheme for general nonlinear systems subject to disturbances. The proposed controller can accept arbitrary forms of dynamical disturbances and carry out risk-aware control by evaluating the CVaR values using Monte-Carlo sampling. Our simulations show that the RA-MPPI controller can effectively reduce the chances of catastrophic scenarios without compromising the overall control performance, even in time-critical applications such as aggressive autonomous driving.  

In the future, we will integrate chance constraints \cite{CSSMPC} and time-consistent modifications of CVaR \cite{Samuelson2018SafetyAwareOC, MarcoPavoneCVaR} with the RA-MPPI control architecture.  
Since different risk metrics interpret risks from their own unique aspect, it would be interesting to compare experimentally these risk-aware controllers. 
In addition, we can further improve the RA-MPPI control architecture by applying ideas similar to \cite{CCMPPI} to achieve adjustable trajectory sampling distributions, hence increasing the sampling efficiency of the proposed algorithm.

\newpage
\bibliographystyle{IEEEtran}
\bibliography{bib/references}

\end{document}